\documentclass{midl}
\graphicspath{{./fig/}}

\jmlryear{2021}
\jmlrworkshop{Full Paper -- MIDL 2021}

\title{An MRF-UNet Product of Experts for Image Segmentation}

\midlauthor{\Name{Mikael Brudfors\nametag{$^{1}$}}
\Email{mikael.brudfors@kcl.ac.uk }\\ \Name{Ya\"el Balbastre\nametag{$^{2}$}}
\Email{ybalbastre@mgh.harvard.edu}\\ \Name{John Ashburner\nametag{$^{3}$}}
\Email{j.ashburner@ucl.ac.uk}\\ \Name{Geraint Rees\nametag{$^{4}$}}
\Email{g.rees@ucl.ac.uk}\\ \Name{Parashkev Nachev\nametag{$^{5}$}}
\Email{p.nachev@ucl.ac.uk}\\ \Name{S\'ebastien Ourselin\nametag{$^{1}$}}
\Email{sebastien.ourselin@kcl.ac.uk}\\ \Name{M. Jorge Cardoso\nametag{$^{1}$}}
\Email{m.jorge.cardoso@kcl.ac.uk}\\ \addr $^{1}$ School of Biomedical
Engineering \& Imaging Sciences, KCL, London, UK \AND \addr $^{2}$ Athinoula A.
Martinos Center for Biomedical Imaging, MGH and HMS, Boston, USA \AND \addr
$^{3}$ Wellcome Center for Human Neuroimaging, UCL, London, UK \AND \addr $^{4}$
Institute of Cognitive Neuroscience, UCL, London, UK \AND \addr $^{5}$ Institute
of Neurology, UCL, London, UK}

\begin{document}

\maketitle

\begin{abstract}

While convolutional neural networks (CNNs) trained by back-propagation have seen
unprecedented success at semantic segmentation tasks, they are known to struggle
on out-of-distribution data. Markov random fields (MRFs) on the other hand,
encode simpler distributions over labels that, although less flexible than
UNets, are less prone to over-fitting. In this paper, we propose to fuse both
strategies by computing the product of distributions of a UNet and an MRF. As
this product is intractable, we solve for an approximate distribution using an
iterative mean-field approach. The resulting MRF-UNet is trained jointly by
back-propagation. Compared to other works using conditional random fields
(CRFs), the MRF has no dependency on the imaging data, which should allow for
less over-fitting. We show on 3D neuroimaging data that this novel network
improves generalisation to out-of-distribution samples. Furthermore, it allows
the overall number of parameters to be reduced while preserving high accuracy.
These results suggest that a classic MRF smoothness prior can allow for less
over-fitting when principally integrated into a CNN model. Our implementation is
available at \url{https://github.com/balbasty/nitorch}.

\end{abstract}


\section{Introduction}

This paper concerns the task of semantic image segmentation: labelling each
voxel of an image with a corresponding class. Robustly identifying voxels of
organs or lesions from medical images is one of the more challenging tasks in
medical image analysis. In this domain, magnetic resonance imaging (MRI) is an
extremely versatile modality, as a variety of image contrasts can be obtained by
changing the multitude of parameters encoding the MR sequence. However, the
resulting image is extremely sensitive to both scanner- and subject-specific
parameters (\emph{e.g.}, field strength, homogeneity of the different magnetic
fields, loading of the coils). This makes building generic segmentation tools --
that work on any contrast or resolution -- extremely challenging due to the
domain shift introduced by both subject and scanner variability.

While classical probabilistic methods, which model the different sources of
artefacts and optimise parameters on a subject-wise basis, generally work well
on out-of-distribution data \cite{ashburner2005unified, zhang2001segmentation,
fischl2004sequence, van1999automated}, neural networks (NNs) have so far
struggled to generalise in the same way \cite{dolz20183d}. This issue is
intrinsically linked to the flexibility of NNs, which makes them extremely good
at recognising patterns but also blind to invariances that are not present in
the data they were trained on. The first methods that tackled generalisation
issues in NN segmentation therefore aimed to pre-process the images to
standardise their intensity profiles \cite{zhuge2009intensity,
weisenfeld2004normalization, han2007atlas}. However, these techniques do not
make the networks generalise \emph{per se} but merely remove (some) variance
from the data.

More recently, two different paths have been taken to build segmentation
networks that are insensitive to certain image characteristics. The first
approach relies on data augmentation, with the underlying idea that, for NNs to
be invariant to some feature, they need this invariance to be discoverable from
the training data. The idea is that the feature (\emph{e.g.}, intensity
non-uniformities) can be modelled, and therefore sampled. Spatial augmentation,
for example, was quickly adopted to present NNs with many more brain shapes than
would be possible using real images alone \cite{pereira2016brain,
castro2018elastic}. As for appearance augmentation, \citet{jog2019psacnn}
generated realistic images with a variety of contrasts using a pulse-sequence
simulator and used them to train a UNet on multiple contrasts. In
\citet{billot2020learning}, the idea was pushed further by generating a
multitude of MR contrasts from pre-segmented MRIs. Importantly, these
simulations did not aim to generate realistic images, but to build contrast
invariance in the training set. More work has since been extended to build
invariance to resolution \cite{billot2020partial}, or even to image features
entirely \cite{hoffmann2020learning}. When labelled images are scarce,
augmentation can be used in conjunction with a consistency loss in a
semi-supervised setting to enforce consistency between predictions obtained from
the same original images augmented in different ways \cite{xie2019unsupervised}.
It has also recently been proposed to add spatial regularization, such as total
variation, to the segmented object and solve by gradient decent
\cite{jia2021regularized}. The second approach focuses on the architecture of
the NNs, such that invariance is directly built-in, independent of the training
data. For example, by adding a new set of batch normalisation parameters in the
network as it encounters training data from a new acquisition protocol
\cite{karani2018lifelong}. Adversarial techniques can also be used to learn
feature representations agnostic to the data domain. This can be achieved by
learning an adversarial network that attempts to discriminate the domain of the
input data coming from both domains \cite{kamnitsas2017unsupervised}. Finally,
transfer learning is one more popular method for improving the generalisability
of NNs \cite{knoll2019assessment}.

A modelling paradigm that can be used to introduce both augmentation and
architectural components is based on probability theory
\cite{jaynes2003probability}. In probabilistic models, the joint distribution
over all variables (observed and hidden) is factorised in a way that reveals
components that influence each observed sample, and components that embed
general knowledge, independent of a particular sample. These components are
commonly denoted as likelihood and prior, respectively. Conversely, most NNs
compute a function that map observed data and hence cannot separate prior and
data components. One line of research aims to bridge the gap between classical
probabilistic models and NNs. In \citet{brudfors2019nonlinear}, it was shown
that inference under a low parameter Markov random field (MRF) prior could be
formalised as a feed-forward NN. The concept was used to encode complex
non-linear MRFs that cannot be optimised in a classical maximum-likelihood
framework, but can be optimised by back-propagation. This MRF was then used to
simply post-process segmentations obtained from a probabilistic segmentation
model.

In the present work, this idea is extended so that a UNet is used in place of
the probabilistic model. In this context, the UNet and MRF are considered as
independent segmentation experts (in the sense that they define probability
distributions over possible segmentations), whose beliefs should be merged in
order to take an informed and balanced decision. Here, this `belief fusion' is
performed by taking the product of their distributions
\citep{hinton2002training}. As normalising this product is intractable, we use
variational inference to estimate the closest factorised distribution under the
Kullback-Leibler divergence. As in \citet{brudfors2019nonlinear}, we find that
this mean-field inference can be formalised as a recurrent NN that is appended
to the UNet. Finally, we propose to jointly train the UNet and MRF by
back-propagation. Encoding the MRF in the NN has advantages over other works
that use CRFs \cite{zheng2015conditional, chen2015learning,
kamnitsas2017efficient, monteiro2018conditional}, as the MRF is a prior
distribution over the segmentation labels, rather than a conditional
distribution. This property allows the MRF to regularise the segmentation labels
alone, without being influenced by the image data, which could allow for less
over-fitting. Additionally, the performance of CRFs have been shown to not
translate well to the medical imaging domain \cite{monteiro2018conditional}.
Conversely, we show in this paper that the MRF-UNet improves segmentation
accuracy on both in- and out-of-distribution 3D brain MRIs. We also show that it
allows the overall number of CNN parameters to be reduced with higher accuracy
preserved. These results suggest that combining a classical type of MRF prior
with a highly parametrised segmentation CNN could improve segmentation accuracy
and generalisability.

\section{Methods}

Let us consider the segmentation problem where an observed intensity image
$\mathbf{X} \in \mathbb{R}^{I \times C}$, with $I$ voxels and $C$ channels, is
segmented into $K$ classes. The segmentation can be encoded in a one-hot label
image $\mathbf{Z} \in \left\{0,1\right\}^{I \times K}$. In the supervised
setting, a set of training pairs $\{\mathbf{X}_n, \hat{\mathbf{Z}}_n\}_{n=1}^N$
is available. This set is used to optimise a function $\mathcal{F}(\mathbf{X})$
that predicts a segment from an image. Currently, functions of choice are
convolutional neural networks (CNNs); often some flavour of UNet
\cite{long2015fully,ronneberger2015u}. The CNN parameters are found by
back-propagating gradients from an appropriate loss function.

Segmentation UNets typically end with a softmax activation function, which
ensures that their output, $\boldsymbol{\pi} \in [0, 1]^{I \times K}$, can be
interpreted as probabilities. We can therefore see the network as encoding a
product of (posterior) categorical distributions:
\begin{equation}
p(\mathbf{Z} \mid \mathbf{X}, \mathcal{F}) = \prod_{i=1}^I
\operatorname{Cat}\left(\mathbf{z}_i \mid \mathcal{F}_i(\mathbf{X})\right) =
\prod_{i=1}^I \prod_{k=1}^K {\mathcal{F}_{ik}(\mathbf{X})}^{z_{ik}} ~.
\end{equation}
Here, $\mathcal{F}$ denotes the UNet parameters and $\mathcal{F}(\mathbf{X})$
the result of its forward pass. The subscripts $i$ and $k$ respectively denote
extracting a single voxel and a single class.

On the other hand, an MRF is a joint probability over all voxels, with the
property that the conditional probability of a voxel, given all others, only
depends on a small neighbourhood $\mathcal{N}$:
\begin{equation}
p\left(\mathbf{z}_i \mid \{\mathbf{z}_j\}_{j\neq i}, \mathcal{W}\right) = 
p\left(\mathbf{z}_i \mid \mathbf{z}_{\mathcal{N}_i}, \mathcal{W}\right) ~,
\end{equation}
where $\mathcal{W}$ denotes the MRF weights. We make the assumption that this
neighbourhood is stationary, meaning that it is defined by relative positions
with respect to $i$. We additionally assume that it factorises over its
neighbours and that each factor is a categorical distribution:
\begin{align}
p(\mathbf{z}_i\mid\mathbf{z}_{\mathcal{N}_i}, \mathcal{W})
= \prod_{\delta \in \mathcal{N}_i} \prod_{k=1}^K \prod_{l=1}^K
\left(w_{kl,\delta}\right)^{z_{ik} \cdot z_{i+\delta,l}}~.
\label{eq:pz2}
\end{align}
This paper uses a first-order neighbourhood, but a larger one could also have
been used.

\begin{algorithm2e}[t]
\KwIn{${\bf X}$, ${\bf R}$ \quad (image data, initial responsibilities)}
\KwOut{${\bf R}^\star$ \quad (VB optimal responsibilities)}
${\bf U} \leftarrow \text{UNet}({\bf X}; \mathcal{F})$\;
${\bf U} \leftarrow {\bf U} - \text{log-sum-exp}({\bf U})$\;
${\bf R} \leftarrow 1/K$\;
\For{$i \leftarrow 1$ \KwTo $niter$}{
${\bf R} \leftarrow \text{softmax}\left({\bf U} + \text{MRF}({\bf R}; 
\mathcal{W})\right)$; \quad (Eq. \eqref{eq:resp})\
}
\caption{MRF-UNet forward pass.}
\label{alg:net}
\end{algorithm2e}

The UNet and MRF distributions can be fused by taking their product and
normalising \cite{hinton2002training}:
\begin{equation}
p\left(\mathbf{Z} \mid \mathbf{X}, \mathcal{F}, \mathcal{W}\right) =
\frac{p\left(\mathbf{Z} \mid \mathcal{W}\right) p\left(\mathbf{Z} \mid
\mathbf{X}, \mathcal{F}\right)} {\int_{\mathbf{Z}} p\left(\mathbf{Z} \mid
\mathcal{W}\right) p\left(\mathbf{Z} \mid \mathbf{X}, \mathcal{F}\right)
\text{d}\mathbf{Z}}~.
\end{equation}
However, the conditional distribution on the left-hand side is clearly
intractable. Instead, we make a mean-field approximation and look for an
approximate distribution $q(\mathbf{Z}) = \prod_i q_i(\vec{z}_i)$, which
factorises across voxels and is closest to the true product of distributions in
terms of their Kullback-Leibler divergence
$\operatorname{KL}\left(q\middle\|p\right)$. As in variational Bayesian
inference, we can update the distribution of a factor by taking the expected
value of the true product of distributions with respect to all the others
factors \cite{bishop2006pattern}:
\begin{align}
\ln q^\star(\mathbf{z}_i) & = \mathbb{E}_{q_{j\neq i}}\left[\ln
p\left(\mathbf{Z} \mid \mathbf{X}, \mathcal{F}, \mathcal{W}\right) \right] +
\text{const} \\ & {}= \sum_{k=1}^K z_{ik} \left(\ln \mathcal{F}_{ik}(\mathbf{X})
+ \sum_{\delta\in\mathcal{N}} \sum_{l=1}^K
\mathbb{E}_q\left[z_{i+\delta,l}\right] \ln w_{kl,\delta}\right) + 
\text{const}~.
\end{align} 
Let us write $\mathbb{E}_q\left[\mathbf{z}_{j}\right] = \mathbf{r}_{j}$. We note
that the second term can be seen as the convolution of the map $\mathbf{R}$ with
a small kernel whose weights are $\ln w_{kl\delta}$ and center weight is zero
\cite{brudfors2019nonlinear}. We denote this convolution $\mathcal{W} \ast
\mathbf{R}$ and we recognise that $q_i^\star$ is a categorical distribution
$\operatorname{Cat}(\mathbf{z}_i \mid \mathbf{r}_i^\star)$ with parameter:
\begin{equation}
\mathbf{r}_{i}^\star = \text{softmax}\left(\ln \mathcal{F}_{i}(\mathbf{X})  +
\left[\mathcal{W} \ast \mathbf{R}\right]_i\right).
\label{eq:resp}
\end{equation}
Note that, by letting the UNet output logits maps (pre-softmax), $\ln
\mathcal{F}_{i}(\mathbf{X})$ can be formulated using the log-sum-exp
trick\footnote{$\log(\text{softmax}({\bf x}))_k = x_k - \log(\sum_j^K\exp(x_j))
= x_k - (x^\star + \log \left[ \sum_j^K \exp \left( x_j - x^\star\right)
\right])$, where $x^\star = \text{max}({\bf x})$.}. The expression in
\eqref{eq:resp} gives us the optimal expected label image, as for the
categorical distribution we have $\mathbb{E}_{q^\ast}[z_{ik}] = r_{ik}^\star$.

\begin{figure*}[t]
\centering
\includegraphics[width=0.8\textwidth]{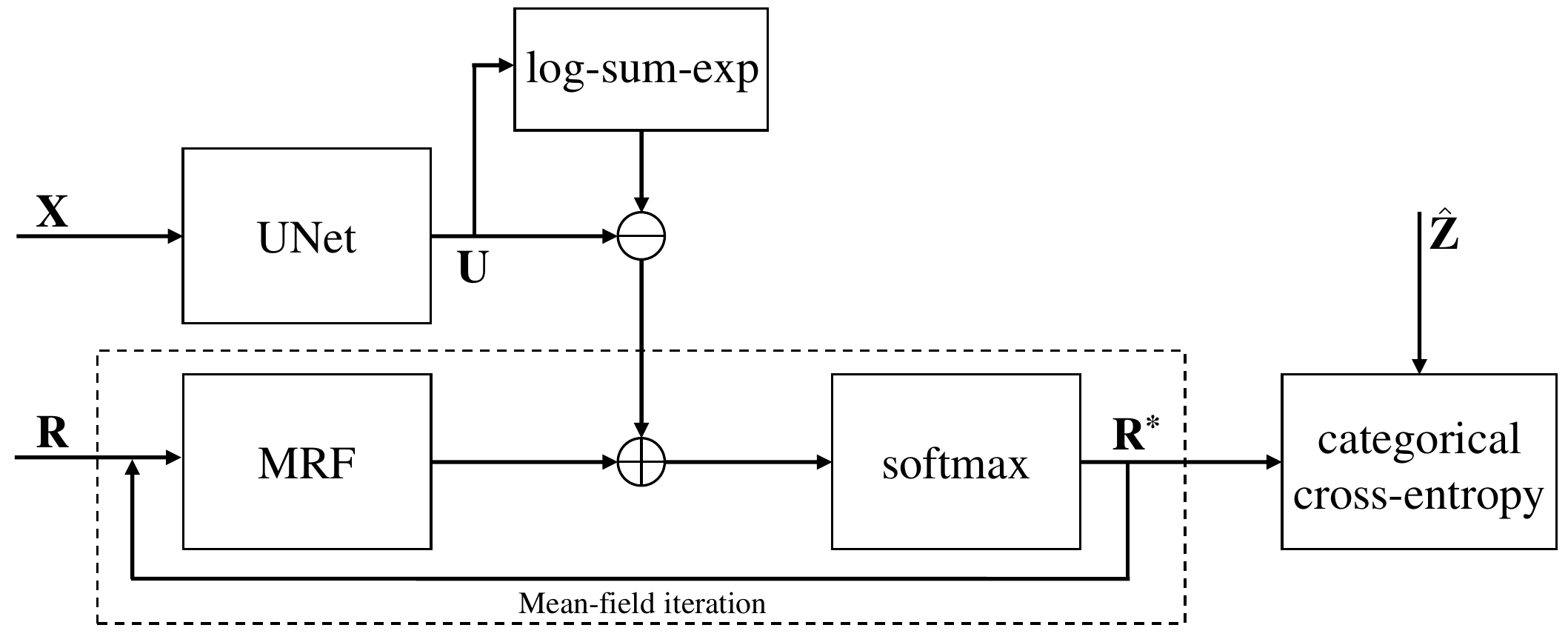}
\caption{Schematic illustration of the MRF-UNet product. An image ${\bf X}$ is
passed forward, through a UNet, whose logit outputs are then fused with the
current estimate of the responsibility map ${\bf R}$. The responsibility map is
updated in an iterative fashion ${\bf R}^\star$. For training, the categorical
cross entropy between the reference segmentation $\hat{{\bf Z}}$ and the
responsibilities is computed.}
\label{fig:mrfnet}
\end{figure*}

In a variational setting, optimising for a segmentation ${\bf R}^\star$ involves
iterating over the expression in \eqref{eq:resp}, which minimises the KL
divergence. To discourage the CNN from over-fitting to a fixed number of
iterations we randomly sample the number of iterations from a discrete uniform
distribution during training, but keep this number fixed during testing. The
MRF-UNet forward pass is described in Algorithm \ref{alg:net} and also
visualised in Figure \ref{fig:mrfnet}.

\section{Validation}

In this section we compare the proposed MRF-UNet architecture to a baseline
model (\emph{i.e.}, a MRF-UNet without the MRF component), for segmenting
publicly available 3D MRI brain scans. We compare the segmentation accuracy of
the two methods on in- and out-of-distribution test data, and how it depends on
the number of network parameters. We additionally investigate the iterative
nature of the MRF-UNet.

\subsection{Data}

\begin{figure*}[t]
\centering
\includegraphics[width=\textwidth]{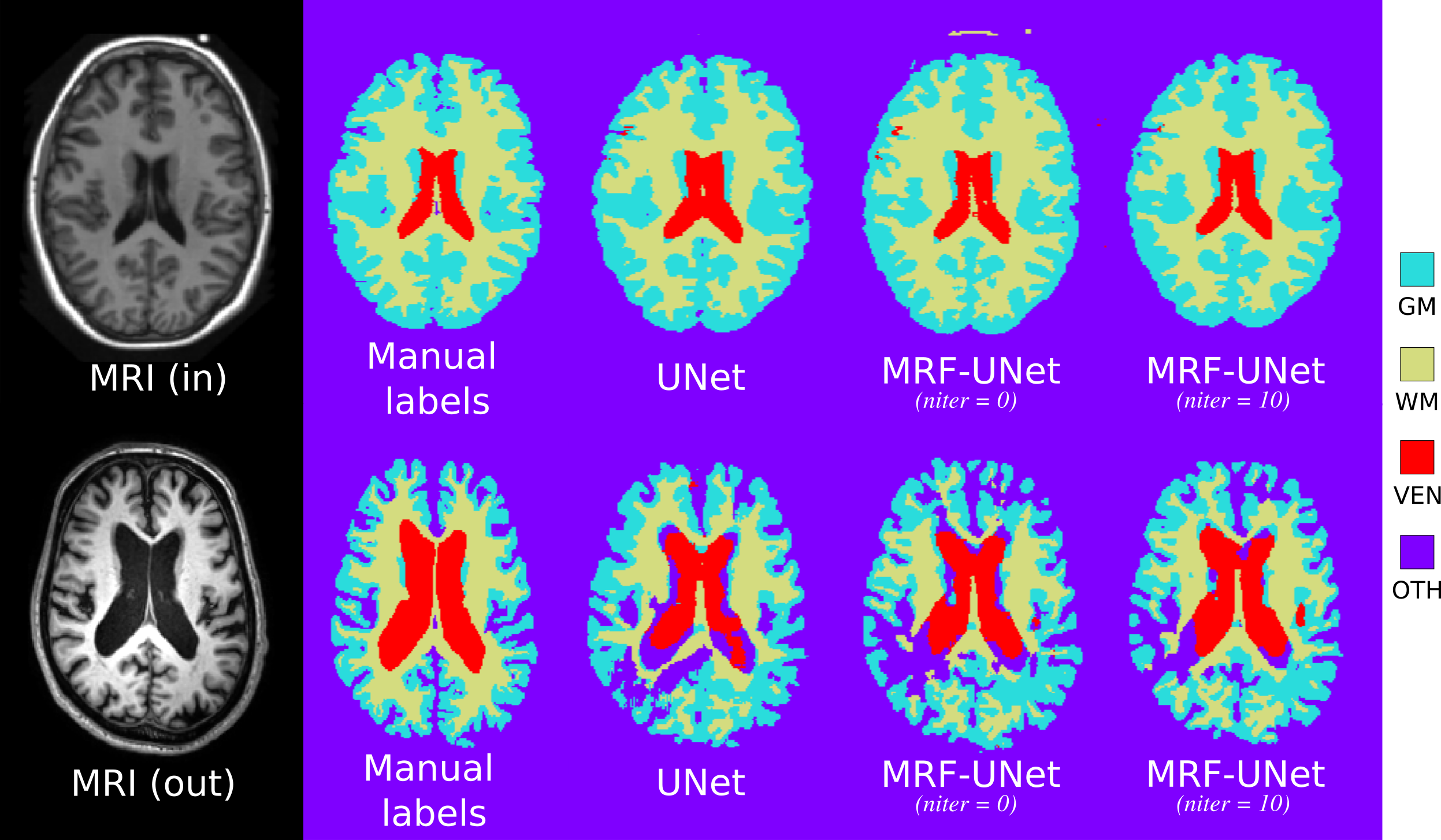}
\caption{Random example segmentation results for in- (top) and
out-of-distribution (bottom) data, as axial slices. The MRI, with its
ground-truth, manual label image, has been segmented either with a UNet or an
MRF-UNet ($j=3$). For the MRF-UNet we show results for zero (without applying
the MRF) and 10 mean-field iterations. A reason that the MRF-UNet ($niter=0$)
segmentation differs from the UNet could be that training a combined MRF-UNet
allows the UNet part to focus on certain image features, as the MRF models some
of the spatial regularity. Therefore, when one does a forward pass though a
trained MRF-UNet, without any mean-field iterations, some of the spatial
smoothness is missing in the segmentation.}
\label{fig:mri-examples}
\end{figure*}

The following MR images from two publicly available datasets are used:
\begin{itemize}
\item 
\textbf{MICCAI2012}\footnote{\url{https://my.vanderbilt.edu/masi/workshops/}}:
T1-weighted MRIs of 30 healthy subjects aged 18 to 96 years, (mean: 34, median:
25). The scans were manually segmented into 136 anatomical regions (by
Neuromorphometrics Inc.) for the MICCAI 2012 multi-atlas segmentation challenge.
We combined regions to form four labels: gray matter (GM), white matter (WM),
ventricles (VEN) and other (OTH).

\item \textbf{MRBrainS18}\footnote{\url{https://mrbrains18.isi.uu.nl/}}:
T1-weighted MRIs of seven subjects all aged 50 years or older (some with
pathology). The scans were manually segmented into ten anatomical regions by the
same neuroanatomist. From these regions we selected the GM, WM, VEN and OTH
labels, for parity with MICCAI2012.
\end{itemize}
Within each dataset, all subjects were imaged on the same scanner and with the
same sequences, whilst between datasets, the scanners and sequences differ. An 
example subject, from both datasets, is shown in Figure \ref{fig:mri-examples}.

\subsection{Implementation}

The UNet has five encoding/decoding layers and use convolutional filters with $3
\times 3 \times 3$ kernels and stride of two. In our experiments, we vary the
number of filters for the encoding layers are  (which are `mirrored' for the
decoding layer). These layers are followed by a final convolution layer that
outputs $K$ channels. The MRF network has an initial $3 \times 3 \times 3$ MRF
layer, whose centre weights are zero with $K^2$ filters, this is followed by one
$1 \times 1 \times 1$ convolution with $K$ filters. The baseline UNet ends with
a softmax activation function, whereas there is no final activation in the UNet
component (nor the MRF component) of the MRF-UNet product. Instead, their logits
are summed before being softmaxed, as depicted in Figure \ref{fig:mrfnet}. All
layers in both networks use leaky ReLU activations ($\alpha=0.2$). The networks
are optimised using categorical cross-entropy and the ADAM optimiser
(lr=$10^{-3}$), where the learning rate is dynamically reduced based on the
difference in subsequent values of the validation loss. During training, we
augment with random diffeomorphic deformations, multiplicative smooth intensity
non-uniformities, and additive Gaussian noise. The MRF-UNet uses
$n_{\text{iter}}=10$ mean-field iterations during testing and $n_{\text{iter}}
\sim \mathcal{U}\{5,15\}$ during training. We train for a fixed number of 50
epochs, with a batch size of one. Our implementation was done using PyTorch.

\subsection{Experiments}

The MICCAI2012 dataset is used as in-distribution data with a (train,
validation, test) split of $(13,~3,~14)$. All seven MRBrainS18 images is used as
out-of-distribution data and considered solely for testing. Both the UNet and
the MRF-UNet are trained on the MICCAI2012 training set; then, mean pairwise
Dice scores are computed for predicting the GM, WM, VEN and OTH labels on the
MICCAI2012, as well as the MRBrainS18, test subjects. This is then done for a
varying number of UNet parameters: the number of filters in the encoding and
decoding layers are set to $(2^j,~2^{(j + 1)},~2^{(j + 2)},~2^{(j + 3)},~2^{(j +
4)})$ for $j=\{1,2,3,4,5,6\}$ (flipped for the decoding layer). We also perform
a simple convergence analysis, where a trained MRF-UNet model ($j=5$) is fitted
to the in- and out-of-distribution data, varying the number of mean-field
iterations from 0 to 20, and computing the average Dice score.

\subsection{Results}

\begin{figure*}[t]
\centering
\includegraphics[width=\textwidth]{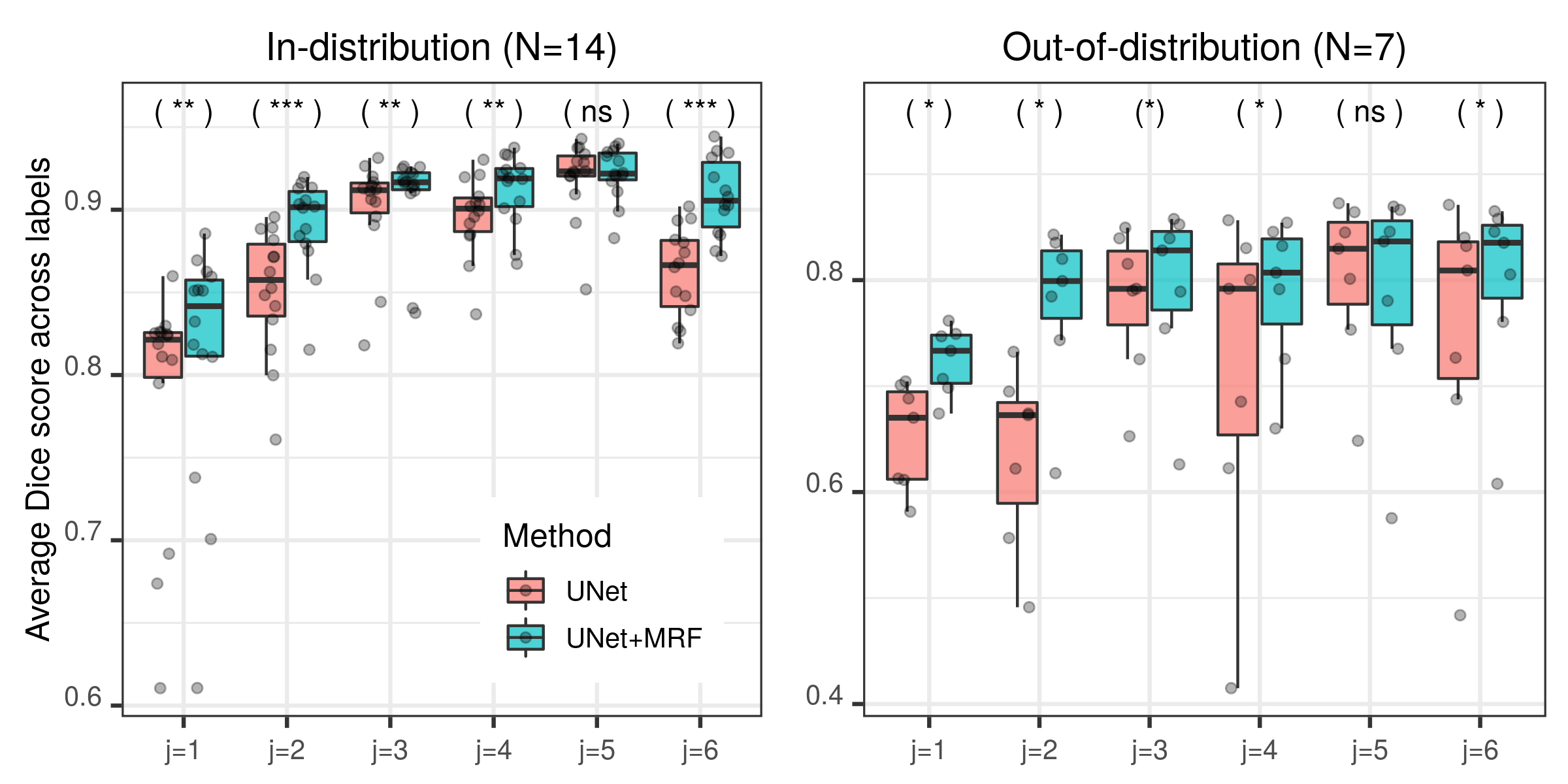}
\caption{Average Dice scores across labels for segmenting the in- and
out-of-distribution test images into GM, WM, VEN and OTH; using both the UNet
and the MRF-UNet. For both networks, we vary the number of convolutional
filters: $(2^j,~2^{(j + 1)},~2^{(j + 2)},~2^{(j + 3)},~2^{(j + 4)})$, for
$j=\{1,2,3,4,5,6\}$. On each box, the central mark indicates the median, and the
bottom and top edges of the box indicate the 25th and 75th percentiles,
respectively. The whiskers extend to the most extreme data points not considered
outliers. The asterisks above the boxes indicate statistical significance of
paired Wilcoxon tests after Bonferroni correction: 0.05 ($\ast$), 0.01
($\ast\ast$), 0.001 ($\ast\ast\ast$).}
\label{fig:mri-dice}
\end{figure*}

\begin{figure*}[t]
\centering
\includegraphics[width=\textwidth]{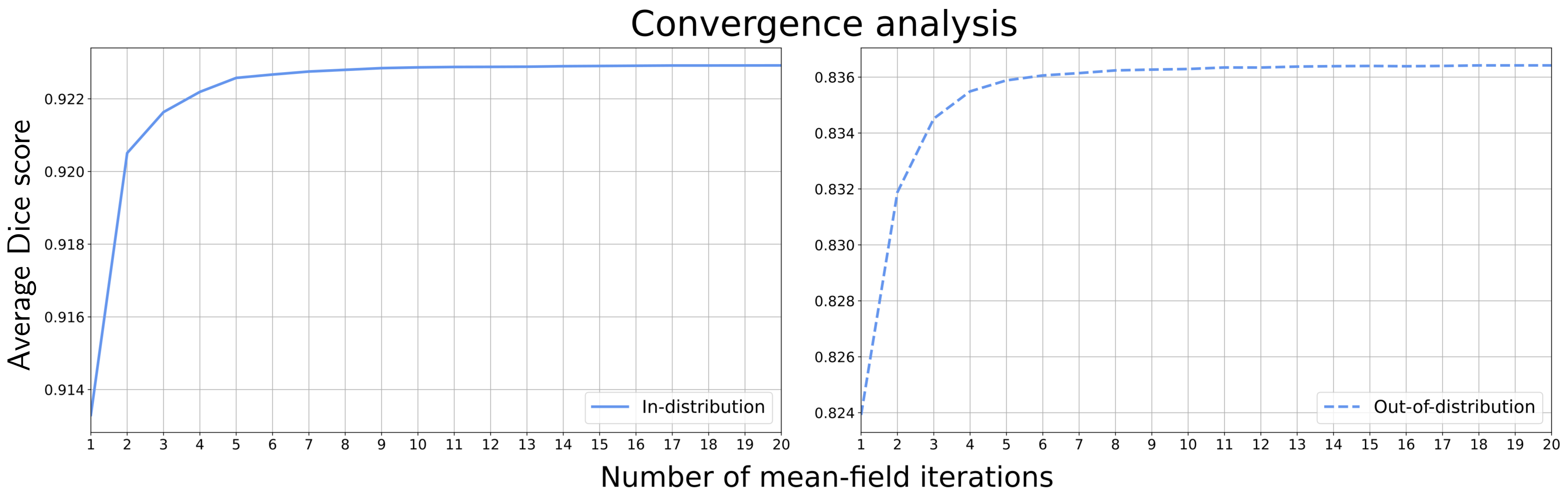}
\caption{Convergence analysis of the learned mean-field iterative approach, on
the in- (left) and out-of-distribution (right) test images. Average Dice scores
were computed for an increasing number of iterations.}
\label{fig:iter}
\end{figure*}

Figure \ref{fig:mri-dice} shows the resulting Dice scores from the in- and
out-of-distribution segmentation tasks. The median Dice scores across labels,
for each parameter configuration, were for the in-distribution task (UNet+MRF
vs. UNet): 0.84 vs 0.82 for $j=1$, 0.90 vs 0.86 for $j=2$, 0.92 vs 0.91 for
$j=3$, 0.92 vs 0.90 for $j=4$, 0.92 vs 0.92 for $j=5$, 0.91 vs 0.87 for $j=6$;
and for the out-of-distribution task: 0.73 vs 0.67 for $j=1$, 0.80 vs 0.67 for
$j=2$, 0.83 vs 0.79 for $j=3$, 0.81 vs 0.79 for $j=4$, 0.84 vs 0.83 for $j=5$,
0.84 vs 0.81 for $j=6$. Paired Wilcoxon tests with Bonferroni correction show
that the segmentation results for $j=\{1,2,3,4,5,6\}$ are significant, except
for $j=5$, for both datasets. That is, the MRF-UNet outperforms the baseline
UNet for almost all parameter configurations. Furthermore, the plot implies that
the MRF-UNet model allows for using fewer UNet parameters, with retained Dice
scores. Introducing the MRF adds parameters to the MRF-UNet, which could results
in a better fit; however, even for the smallest architecture considered in our
experiments, $(2,~4,~8,~16,~32)$, the increase in parameters is less than
$2.5\%$. For the largest architecture, this drops to less than $0.0001\%$, which
shows how lightweight the MRF component is. Figure \ref{fig:mri-examples} shows
example segmentations for both datasets. Segmenting out-of-distribution images
is clearly a very challenging task, having only seen the in-distribution data.
However, it can be seen, from comparing the MRF-UNet with 0 mean-field
iterations (no MRF applied) to 10 iterations, that the MRF component behaves as
expected, encouraging neighbouring voxels to have similar labels. Figure
\ref{fig:iter} shows the results of the convergence analysis. For both the in-
and out-of-distribution data, the validation Dice converges quickly and
monotonically. The analysis suggests that no more than ten iterations may be
needed, which is beneficial from both a memory and a runtime point-of-view. It
is furthermore encouraging that the learned mean-field iterative approach
replicates the monotonically increasing nature of variational updates.

\section{Conclusion}

In this paper, we described a novel approach for combining a segmentation CNN
with a low parameter, first-order MRF prior over the image labels. Our
hypothesis was that this `simple' prior would learn abstract, label-specific
features and thereby improve segmentation accuracy on both in- and
out-of-distribution data. We showed the validity of this assumption on 3D MR
images of the human brain. Future work will extend this validation to data from
other domains. One could argue that explicitly encoding prior information into a
high-dimensional model, such as a CNN, is superfluous, as the CNN should
implicitly capture this information from training data. However, in the interest
of limited data, and model complexity, explicit priors still play an important
role.

Readers familiar with unsupervised segmentation techniques may notice that the
expression for updating the prediction of a segmentation in \eqref{eq:resp},
coincides with updating the expected posterior over latent segmentation labels,
where the likelihood is a mixture model and the prior an MRF
\cite{langan1992use, van1999automated}. In this work, it is not possible to
encode the posterior using Bayes' rule as the UNet outputs a conditional
distribution over segmentation labels, not imaging data; and we here chose to
instead use a products of experts model. However, the connection between the two
methods is clear and could inspire future extensions of our approach.

\midlacknowledgments{MB, PN and MJC were supported by Wellcome Innovations
[WT213038/Z/18/Z]. PN  was  supported  by the  UCLH  NIHR Biomedical  Research
Centre. YB was supported by  the  National  Institutes  of  Health  under award 
numbers U01MH117023, R01AG064027 and P41EB030006.}

\bibliography{bibliography}

\end{document}